\documentclass[sigconf]{acmart}
\renewcommand\footnotetextcopyrightpermission[1]{} % removes footnote with conference information in first column
\usepackage{microtype}
\usepackage{graphicx}
\usepackage{subcaption}
\usepackage{booktabs} % for professional tables
\usepackage{amsmath}
\usepackage{amssymb}
\usepackage{mathtools}
\usepackage{enumitem}
\usepackage{hyperref}
\usepackage{lipsum}
\usepackage{algorithm}
\usepackage{algorithmic}

\usepackage{etoolbox}

\DeclareMathOperator*{\argmin}{argmin}
\DeclarePairedDelimiter\abs{\lvert}{\rvert}

\def\BibTeX{{\rm B\kern-.05em{\sc i\kern-.025em b}\kern-.08emT\kern-.1667em\lower.7ex\hbox{E}\kern-.125emX}}

\setcopyright{none}

\begin{document}

%
% The "title" command has an optional parameter, allowing the author to define a "short title" to be used in page headers.
\title{Auto-Keras: An Efficient Neural Architecture Search System}

\author{Haifeng Jin, Qingquan Song, Xia Hu}
\affiliation{
  \institution{Department of Computer Science and Engineering, Texas A\&M University} %  77843}
  % \city{$^\mathsmaller{\ddagger}$Center for Remote Health Technologies and Systems, Texas A\&M Engineering Experiment Station} %  77843
  }
 \email{{jin,song_3134,xiahu}@tamu.edu}

\renewcommand{\shortauthors}{H. Jin et al.}

%
% The "author" command and its associated commands are used to define the authors and their affiliations.
% Of note is the shared affiliation of the first two authors, and the "author note" and "authornotemark" commands
% used to denote shared contribution to the research.

% \author{Haifeng Jin}
% \affiliation{%
%   \institution{Department of Computer Science and Engineering, Texas A\&M University}
%   \city{College Station}
% \email{jin@tamu.edu}
%
% \author{Qingquan Song}
% \affiliation{%
%   \institution{Department of Computer Science and Engineering, Texas A\&M University}
%   \city{College Station}
% \email{song_3134@tamu.edu}
%
% \author{Xia Hu}
% \affiliation{%
%   \institution{Department of Computer Science and Engineering, Texas A\&M University}
%   \city{College Station}
% \email{xiahu@tamu.edu}
%
%
% By default, the full list of authors will be used in the page headers. Often, this list is too long, and will overlap
% other information printed in the page headers. This command allows the author to define a more concise list
% of authors' names for this purpose.

%
% The abstract is a short summary of the work to be presented in the article.
\begin{abstract}
Neural architecture search (NAS) has been proposed to automatically tune deep neural networks, but existing search algorithms, \textit{e.g.}, NASNet~\cite{zoph2016neural}, PNAS~\cite{liu2017progressive}, usually suffer from expensive computational cost. Network morphism, which keeps the functionality of a neural network while changing its neural architecture, could be helpful for NAS by enabling more efficient training during the search. In this paper, we propose a novel framework enabling Bayesian optimization to guide the network morphism for efficient neural architecture search. The framework develops a neural network kernel and a tree-structured acquisition function optimization algorithm to efficiently explores the search space. Intensive experiments on real-world benchmark datasets have been done to demonstrate the superior performance of the developed framework over the state-of-the-art methods. Moreover, we build an open-source AutoML system based on our method, namely Auto-Keras.\footnote{The code and documentation are available at \textcolor{blue}{\url{https://autokeras.com}}} The system runs in parallel on CPU and GPU, with an adaptive search strategy for different GPU memory limits.
\end{abstract}

% Keywords. The author(s) should pick words that accurately describe the work being
% presented. Separate the keywords with commas.

%
% A "teaser" image appears between the author and affiliation information and the body
% of the document, and typically spans the page.
%   \Description{Enjoying the baseball game from the third-base seats. Ichiro Suzuki preparing to bat.}

%
% This command processes the author and affiliation and title information and builds
% in the first part of the formatted document.
\maketitle

\section{Introduction}
Automated Machine Learning (AutoML) has become a very important research topic with wide applications of machine learning techniques.
The goal of AutoML is to enable people with limited machine learning background knowledge to use the machine learning models easily.
Work has been done on automated model selection, automated hyperparameter tunning, and etc.
In the context of deep learning, neural architecture search (NAS), which aims to search for the best neural network architecture for the given learning task and dataset, has become an effective computational tool in AutoML.
Unfortunately, existing NAS algorithms are usually computationally expensive.
The time complexity of NAS is $O(n\bar{t})$, where $n$ is the number of neural architectures evaluated during the search, and $\bar{t}$ is the average time consumption for evaluating each of the $n$ neural networks.
Many NAS approaches, such as deep reinforcement learning~\cite{zoph2016neural,baker2016designing,zhong2017practical,pham2018efficient}, gradient-based methods~\cite{luo2018neural} and evolutionary algorithms~\cite{real2017large,desell2017large,liu2017hierarchical,suganuma2017genetic,real2018regularized}, require a large $n$ to reach a good performance. Also, each of the $n$ neural networks is trained from scratch which is very slow.
% Another branch of work use
%More about the defects of ENAS and DARTS.
% Initial attempts have been made to accelerate the search by predicting the performance of the neural networks instead of actually training them \cite{brock2017smash,zhouneural}. However, they use a deep model as the surrogate model for performance prediction or generating the weights for a neural network, which still requires a large number of trained neural architectures.

Initial efforts have been devoted to making use of network morphism in neural architecture search~\cite{cai2018efficient,elsken2017simple}.
It is a technique to morph the architecture of a neural network but keep its functionality~\cite{chen2015net2net,wei2016network}.
Therefore, we are able to modify a trained neural network into a new architecture using the network morphism operations,
\textit{e.g.}, inserting a layer or adding a skip-connection.
Only a few more epochs are required to further train the new architecture towards better performance.
Using network morphism would reduce the average training time $\bar{t}$ in neural architecture search.
The most important problem to solve for network morphism-based NAS methods is the selection of operations,
which is to select an operation from the network morphism operation set to morph an existing architecture to a new one.
The network morphism-based NAS methods are not efficient enough.
They either require a large number of training examples~\cite{cai2018efficient},
or inefficient in exploring the large search space~\cite{elsken2017simple}.
How to perform efficient neural architecture search with network morphism remains a challenging problem.

As we know, Bayesian optimization~\cite{snoek2012practical} has been widely adopted to efficiently explore black-box functions for global optimization, whose observations are expensive to obtain.
For example, it has been used in hyperparameter tuning for machine learning models~\cite{thornton2013auto,kotthoff2016auto,feurer2015efficient,hutter2011sequential},
in which Bayesian optimization searches among different combinations of hyperparameters.
During the search, each evaluation of a combination of hyperparameters involves an expensive process of training and testing the machine learning model,
which is very similar to the NAS problem.
The unique properties of Bayesian optimization motivate us to explore its capability in guiding the network morphism
to reduce the number of trained neural networks $n$ to make the search more efficient.

It is non-trivial to design a Bayesian optimization method for network morphism-based NAS due to the following challenges.
First, the underlying Gaussian process (GP) is traditionally used for learning probability distribution of functions in Euclidean space.
To update the Bayesian optimization model with observations,
the underlying GP is to be trained with the searched architectures and their performances.
However, the neural network architectures are not in Euclidean space and hard to parameterize into a fixed-length vector.
Second,
an acquisition function needs to be optimized for Bayesian optimization to generate the next architecture to observe.
However, in the context of network morphism, it is not to maximize a function in Euclidean space,
but finding a node in a tree-structured search space,
where each node represents a neural architecture and each edge is a morph operation.
Thus traditional gradient-based methods cannot be simply applied.
Third, the changes caused by a network morphism operation is complicated.
A network morphism operation on one layer may change the shapes of some intermediate output tensors,
which no longer match input shape requirements of the layers taking them as input.
How to maintain such consistency is a challenging problem.

In this paper, an efficient neural architecture search with network morphism is proposed,
which utilizes Bayesian optimization to guide through the search space by selecting the most promising operations each time.
To tackle the aforementioned challenges, an edit-distance neural network kernel is constructed.
Being consistent with the key idea of network morphism, it measures how many operations are needed to change one neural network to another.
Besides, a novel acquisition function optimizer, which is capable of balancing between the exploration and exploitation,
is designed specially for the tree-structure search space
to enable Bayesian optimization to select from the operations.
In addition, a graph-level network morphism is defined to address the changes in the neural architectures based on layer-level network morphism.
The proposed approach is compared with the state-of-the-art NAS methods~\cite{kandasamy2018neural,elsken2017simple} on benchmark datasets of MNIST, CIFAR10, and FASHION-MNIST.
Within a limited search time,
the architectures found by our method achieves the lowest error rates on all of the datasets.

In addition, we have developed a widely adopted open-source AutoML system based on our proposed method,
namely Auto-Keras.
% There are some existing AutoML systems available nowadays.
% Auto-sklearn~\cite{feurer2015efficient}, Auto-WEKA~\cite{thornton2013auto}, TPOT~\cite{OlsonGECCO2016}
% can generate machine learning pipelines for tabular dataset,
% which are fast and easy-to-use.
% For data types need deep learning models, \textit{e.g.} image and text,
% there are cloud-based AutoML systems.
% However, they are usually non-free, requires the user to use Kubernetes and Docker containers to connect.
It is an open-source AutoML system, which can be download and installed locally.
The system is carefully designed with a concise interface for people not specialized in computer programming and data science to use.
To speed up the search, the workload on CPU and GPU can run in parallel.
To address the issue of different GPU memory, which limits the size of the neural architectures, a memory adaption strategy is designed for deployment.

The main contributions of the paper are as follows:
\begin{itemize}[leftmargin=*]
\item Propose an algorithm for efficient neural architecture search based on network morphism guided by Bayesian optimization.
% \item Propose a neural network kernel for Bayesian optimization, a tree-structured acquisition function optimizer, and a graph-level morphism.
\item Conduct intensive experiments on benchmark datasets to demonstrate the superior performance of the proposed method over the baseline methods.
% \item Develop an open-source AutoML system, namely Auto-Keras, based on our method for neural architecture search.
% \item A graph-level network morphism used in neural architecture search is formally defined.
\item Develop an open-source system, namely Auto-Keras, which is one of the most widely used AutoML systems.
\end{itemize}

\section{Problem Statement}
\label{sec_problem}

The general neural architecture search problem we studied in this paper is defined as:  Given a neural architecture search space $\mathcal{F}$,
the input data $D$ divided into $D_{train}$ and $D_{val}$,
and the cost function $Cost(\cdot)$,
we aim at finding an optimal neural network $f^* \in \mathcal{F} $, which could achieve the lowest cost on dataset $D$.
The definition is equivalent to finding $f^*$ satisfying:
\begin{equation}%\small
\label{eq_obj}
f^*=\argmin_{f \in \mathcal{F}} Cost(f(\boldsymbol{\theta}^*), D_{val}),
\end{equation}
\begin{equation}%\small
  \label{eq_theta}
\boldsymbol{\theta}^*=\argmin_{\boldsymbol{\theta}}\mathcal{L}(f(\boldsymbol{\theta}), D_{train}).
\end{equation}
% To obtain the value of $Cost(f, D)$, $D$ is divied into training set
% $D_t=(\boldsymbol{X}_t, \boldsymbol{y}_t)$
% and validation set $D_v=(\boldsymbol{X}_v, \boldsymbol{y}_v)$.
% $Cost(f, D)=Metric(f(\boldsymbol{X_v}, \boldsymbol{\theta}^*), y_v)$,
where $Cost(\cdot, \cdot)$ is the evaluation metric function, \textit{e.g.}, accuracy, mean sqaured error,
$\boldsymbol{\theta}^*$ is the learned parameter of $f$.

% Before explaining the proposed algorithm, we first define the target search space $\mathcal{F}$. Let $G_f=(V_f, E_f)$ denotes the computational graph of a neural network $f$. Each node $v \in V_f$ denotes an intermediate output tensor of a layer of $f$. Each directed edge $e_{u\rightarrow v} \in E_f$ denotes a layer of $f$, whose input tensor is $u \in V_f$ and output tensor is $v\in V_f$. $u \prec v$ indicates $v$ is before $u$ in topological order of the nodes, \textit{i.e.}, by traveling through the edges in $E_f$, $u$ is reachable from node $v$. The search space $\mathcal{F}$
% is defined as:
The search space
% depends on the initial architectures for the search.
% $\mathcal{F}$ is a subspace of feed-forward neural architectures,
% whose computational graph can be represented by a directed acyclic graph (DAG).
$\mathcal{F}$ covers all the neural architectures, which can be morphed from the initial architectures.
The details of the morph operations are introduced in~\ref{sec_nm}.
Notably, the operations can change the number of filters in a convolutional layer,
which makes $\mathcal{F}$ larger than methods with fixed layer width~\cite{liu2018darts}.

\section{Network Morphism Guided by Bayesian Optimization}
\label{sec_method}
The key idea of the proposed method is to explore the search space via morphing the neural architectures guided by Bayesian optimization (BO) algorithm. Traditional Bayesian optimization consists of a loop of three steps: update, generation, and observation.
In the context of NAS, our proposed Bayesian optimization algorithm iteratively conducts: (1) \textbf{Update}: train the underlying Gaussian process model with the existing architectures and their performance; (2)  \textbf{Generation}: generate the next architecture to observe by optimizing a delicately defined acquisition function;
(3) \textbf{Observation}: obtain the actual performance by training the generated neural architecture. There are three main challenges in designing a method for morphing the neural architectures with Bayesian optimization.
We introduce three key components separately in the subsequent sections coping with the three challenges.

\begin{figure*}
    \includegraphics[width=0.85\textwidth]{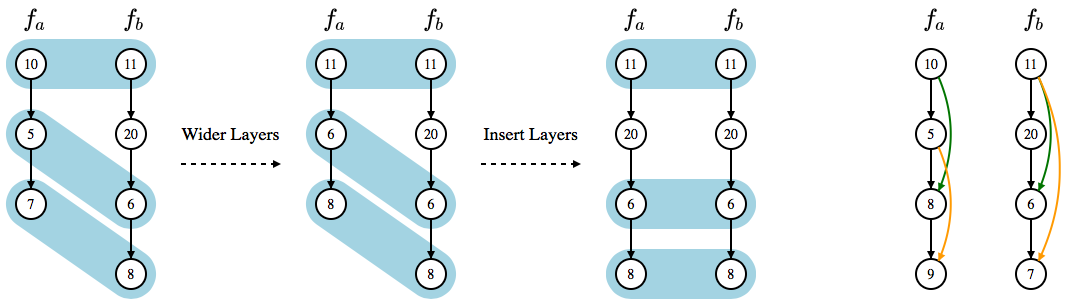}
    \caption{Neural Network Kernel.
    Given two neural networks $f_a$, $f_b$, and matchings between the similar layers, the figure shows how the layers of $f_a$ can be changed to the same as $f_b$. Similarly, the skip-connections in $f_a$ also need to be changed to the same as $f_b$ according to a given matching.}
    \label{fig_layers}
\end{figure*}%

\subsection{Edit-Distance Neural Network Kernel for Gaussian Process}
The first challenge we need to address is that the NAS space is not a Euclidean space, which does not satisfy the assumption of traditional Gaussian process (GP). Directly vectorizing the neural architecture is impractical due to the uncertain number of layers and parameters it may contain. Since the Gaussian process is a kernel method, instead of vectorizing a neural architecture, we propose to tackle the challenge by designing a neural network kernel function. The intuition behind the kernel function is the edit-distance for morphing one neural architecture to another.
More edits needed from one architecture to another means the further distance between them,
thus less similar they are.
The proof of validity of the kernel function is presented in Appendix~\ref{sec_proof}.

\textbf{Kernel Definition}:  Suppose $f_a$ and $f_b$ are two neural networks. Inspired by Deep Graph Kernels~\cite{yanardag2015deep}, we propose an edit-distance kernel for neural networks.  Edit-distance here means how many operations are needed to morph one neural network to another. The concrete kernel function is defined as:
\begin{equation}
  \label{eq_kernel}
  \kappa(f_a, f_b)=e^{-\rho^2(d(f_a,f_b))},
\end{equation}
where function $d(\cdot,\cdot)$ denotes the edit-distance of two neural networks,
whose range is $[0, +\infty)$,
$\rho$ is a mapping function,
which maps the distance in the original metric space to the corresponding distance in the new space.
The new space is constructed by embedding the original metric space into a new one using Bourgain Theorem~\cite{bourgain1985lipschitz},
which ensures the validity of the kernel.

Calculating the edit-distance of two neural networks can be mapped to calculating the edit-distance of two graphs,
which is an NP-hard problem~\cite{zeng2009comparing}.
Based on the search space $\mathcal{F}$ defined in Section~\ref{sec_problem},
we tackle the problem by proposing an approximated solution as follows:
\begin{equation}
\label{eq_distance}
d(f_a, f_b)=D_l(L_a, L_b) + \lambda D_s(S_a, S_b),
\end{equation}
where $D_l$ denotes the edit-distance for morphing the layers, \textit{i.e.}, the minimum edits needed to morph $f_a$ to $f_b$ if the skip-connections are ignored, $L_a=\{l_a^{(1)},l_a^{(2)},\ldots\}$ and $L_b=\{l_b^{(1)},l_b^{(2)},\ldots\}$ are the layer sets of neural networks $f_a$ and $f_b$, $D_s$ is the approximated edit-distance for morphing skip-connections between two neural networks, $S_a=\{s_a^{(1)},s_a^{(2)},\ldots\}$ and $S_b=\{s_b^{(1)},s_b^{(2)},\ldots\}$ are the skip-connection sets of neural network $f_a$ and $f_b$, and
$\lambda$ is the balancing factor between the distance of the layers and the skip-connections.

\textbf{Calculating $D_l$}: We assume  $|L_a|<|L_b|$, the edit-distance for morphing the layers of two neural architectures $f_a$ and $f_b$ is calculated by minimizing the follow equation:
\begin{equation}
\label{eq_dl}
D_l(L_a, L_b)=\min \sum_{i=1}^{|L_a|} d_l(l_a^{(i)}, \varphi_l(l_a^{(i)})) + \abs[\Big]{|L_b| - |L_a|},
\end{equation}
where $\varphi_l:L_a\rightarrow L_b$ is an injective matching function of layers satisfying: $\forall i < j$, $\varphi_l(l_a^{(i)})\prec \varphi_l(l_a^{(j)})$ if layers in $L_a$ and $L_b$ are all sorted in topological order. $d_l(\cdot,\cdot)$ denotes the edit-distance of widening a layer into another defined in Equation~\eqref{eq_small_dl},
\begin{equation}
\label{eq_small_dl}
d_l(l_a, l_b)=\frac{|w(l_a)-w(l_b)|}{max[w(l_a), w(l_b)]},
\end{equation}
where $w(l)$ is the width of layer $l$.

The intuition of Equation~\eqref{eq_dl} is consistent with the idea of network morphism shown in Figure~\ref{fig_layers}.
Suppose a matching is provided between the nodes in two neural networks.
The sizes of the tensors are indicators of the width of the previous layers
(\textit{e.g.}, the output vector length of a fully-connected layer or the number of filters of a convolutional layer).
The matchings between the nodes are marked by light blue.
So a matching between the nodes can be seen as matching between the layers.
To morph $f_a$ to $f_b$ with the given matching,
we need to first widen the three nodes in $f_a$ to the same width as their matched nodes in $f_b$,
and then insert a new node of width 20 after the first node in $f_a$.
Based on this morphing scheme, the edit-distance of the layers is defined as $D_l$ in Equation~\eqref{eq_dl}.

Since there are many ways to morph $f_a$ to $f_b$, to find the best matching between the nodes that minimizes $D_l$, we propose a dynamic programming approach by defining a matrix $\boldsymbol{A}_{|L_a|\times|L_b|}$, which is recursively calculated as follows:
\begin{equation}
\label{eq_dp}
\boldsymbol{A}_{i,j}=max[\boldsymbol{A}_{i-1,j} + 1, \boldsymbol{A}_{i,j-1} + 1, \boldsymbol{A}_{i-1,j-1} + d_l(l_a^{(i)}, l_b^{(j)})],
\end{equation}
where $\boldsymbol{A}_{i,j}$ is the minimum value of $D_l(L^{(i)}_a, L^{(j)}_b)$,
where $L^{(i)}_a=\{l_a^{(1)},l_a^{(2)},\ldots,l_a^{(i)}\}$ and $L^{(j)}_b=\{l_b^{(1)},l_b^{(2)},\ldots,l_b^{(j)}\}$.

\textbf{Calculating $D_s$}:  The intuition of $D_s$ is the sum of the the edit-distances of the matched skip-connections in two neural networks into pairs. As shown in Figure~\ref{fig_layers}, the skip-connections with the same color are matched pairs. Similar to $D_l(\cdot, \cdot)$, $D_s(\cdot, \cdot)$ is defined as follows:
\begin{equation}
\label{eq_ds}
D_s(S_a, S_b)=\min \sum_{i=1}^{|S_a|} d_s(s_a^{(i)}, \varphi_s(s_a^{(i)})) + \abs[\Big]{|S_b| - |S_a|},
\end{equation}
where we assume $|S_a|<|S_b|$. $(|S_b| - |S_a|)$ measures the total edit-distance for non-matched skip-connections since each of the non-matched skip-connections in $S_b$ calls for an edit of inserting a new skip connection into $f_a$.
The mapping function $\varphi_s:S_a\rightarrow S_b$ is an injective function. $d_s(\cdot,\cdot)$ is the edit-distance for two matched skip-connections defined as:
\begin{equation}
\label{eq_small_ds}
d_s(s_a, s_b)=\frac{|u(s_a)-u(s_b)|+|\delta(s_a)-\delta(s_b)|}{max[u(s_a), u(s_b)]+max[\delta(s_a), \delta(s_b)]},
\end{equation}
where $u(s)$ is the topological rank of the layer the skip-connection $s$ started from,
$\delta(s)$ is the number of layers between the start and end point of the skip-connection $s$.

This minimization problem in Equation~\eqref{eq_ds} can be mapped to a bipartite graph matching problem,
where $f_a$ and $f_b$ are the two disjoint sets of the graph,
each skip-connection is a node in its corresponding set.
The edit-distance between two skip-connections is the weight of the edge between them.
The weighted bipartite graph matching problem is solved by the Hungarian algorithm (Kuhn-Munkres algorithm)~\cite{kuhn1955hungarian}.

% \textbf{Proof of Kernel Validity}: Gaussian process requires the kernel to be valid,
% \textit{i.e.}, the kernel matrices are positive semidefinite,
% to keep the distributions valid.
% The edit-distance in Equation \eqref{eq_distance} is a metric distance proved by Theorem 1.
% Though, a generalized RBF kernel in the form of $e^{-\gamma d(x, y)}$ based on a distance in metric space
% may not always be a valid kernel,
% our kernel defined in Equation~\eqref{eq_kernel} is proved to be valid by Theorem 2.

% \textbf{Theorem 1.} $d(f_a, f_b)$ is a metric space distance.

% \textbf{\textit{Proof of Theorem 1:}}
% See appendix.\hfill$\square$

% \textbf{Theorem 2.} $\kappa(f_a, f_b)$ is a valid kernel.

% \textbf{\textit{Proof of Theorem 2:}}
% The kernel matrix of generalized RBF kernel in the form of $e^{-\gamma D^2(x, y)}$ is positive definite
% if and only if there is an isometric embedding in Euclidean space for the metric space with metric $D$ \cite{haasdonk2004learning}.
% Any finite metric space distance can be isometrically embedded into Euclidean space by changing the scale of the distance measurement \cite{maehara2013euclidean}.
% By using Bourgain theorem \cite{bourgain1985lipschitz},
% metric space $d$ is embedded to Euclidean space with little distortion.
% $\rho(d(f_a, f_b))$ is the embedded distance for $d(f_a, f_b)$.
% Therefore, $e^{-\rho^2(d(f_a, f_b))}$ is always positive definite.
% So $\kappa(f_a, f_b)$ is a valid kernel.\hfill$\square$

\subsection{Optimization for Tree Structured Space}
\label{sec_acq}

The second challenge of using Bayesian optimization to guide network morphism is the optimization of the acquisition function.
The traditional acquisition functions are defined on Euclidean space.
The optimization methods are not applicable to the tree-structured search via network morphism.
% The state-of-the-art acquisition function optimization techniques, \textit{e.g.}, gradient-based or Newton-like method,
% are designed for numerical data,
% which cannot be directly applied in the tree-structure space.
% TreeBO~\cite{jenatton2017bayesian} has proposed a way to maximize the acquisition function in tree-structured parameter space.
% However, only the leaf nodes of its tree have acquisition function values, which is different from our case.
% Moreover, the proposed solution is a surrogate multivariate Bayesian optimization model.
% In NASBOT~\cite{kandasamy2018neural}, they use an evolutionary algorithm to optimize the acquisition function.
% They are both very expansive in computing time.
To optimize our acquisition function, we need a method to efficiently optimize the acquisition function in the tree-structured space.
To deal with this problem, we propose a novel method to optimize the acquisition function on tree-structured space.

Upper-confidence bound (UCB)~\cite{auer2002finite} is selected as our acquisition function,
which is defined as:
\begin{equation}
\label{eq_acq}
\alpha(f) = \mu(y_f) - \beta\sigma(y_f),
\end{equation}
where $y_{f} = Cost(f, D)$,
$\beta$ is the balancing factor,
$\mu(y_f)$ and $\sigma(y_f)$ are the posterior mean and standard deviation of variable $y_f$.
It has two important properties, which fit our problem.
First, it has an explicit balance factor $\beta$ for exploration and exploitation.
Second, $\alpha(f)$ is directly comparable with the cost function value $c^{(i)}$ in search history $\mathcal{H}=\{(f^{(i)}, \boldsymbol{\theta}^{(i)}, c^{(i)})\}$.
It estimates the lowest possible cost given the neural network $f$.
$\hat{f}=argmin_{f}\alpha(f)$ is the generated neural architecture for next observation.

The tree-structured space is defined as follows.
During the optimization of $\alpha(f)$, $\hat{f}$ should be obtained from $f^{(i)}$ and $O$,
where $f^{(i)}$ is an observed architecture in the search history $\mathcal{H}$,
$O$ is a sequence of operations to morph the architecture into a new one.
Morph $f$ to $\hat{f}$ with $O$ is denoted as $\hat{f}\leftarrow \mathcal{M}(f, O)$,
where $\mathcal{M}(\cdot,\cdot)$ is the function to morph $f$ with the operations in $O$.
Therefore, the search can be viewed as a tree-structured search,
where each node is a neural architecture, whose children are morphed from it by network morphism operations.

The most common defect of network morphism is it only grows the size of the architecture instead of shrinking them.
Using network morphism for NAS may end up with a very large architecture without enough exploration on the smaller architectures.
However, our tree-structure search, we not only expand the leaves but also the inner nodes,
which means the smaller architectures found in the early stage can be selected multiple times to morph to more comparatively small architectures.

Inspired by various heuristic search algorithms for exploring the tree-structured search space
and optimization methods balancing between exploration and exploitation,
a new method based on A* search
and simulated annealing is proposed.
A* algorithm is widely used for tree-structure search.
It maintains a priority queue of nodes and keeps expanding the best node in the queue.
Since A* always exploits the best node, simulated annealing is introduced to balance the exploration and exploitation by not selecting the estimated best architecture with a probability.

\begin{algorithm}
\caption{Optimize Acquisition Function}\label{algo_max}
\begin{algorithmic}[1]
   \STATE {\bfseries Input:} $\mathcal{H}$, $r$, $T_{low}$
   % \STATE {\bfseries Output:} $f$, $O$
\STATE $T\leftarrow 1$, $Q\leftarrow PriorityQueue()$\
\STATE $c_{min}\leftarrow$ lowest $c$ in $\mathcal{H}$\
\FOR{$(f,\boldsymbol{\theta_f},c) \in \mathcal{H}$}
      \STATE $Q.Push(f)$\
\ENDFOR
\WHILE {$Q\neq\varnothing$ {\bf and} $T > T_{low}$ }
  \STATE $T\leftarrow T\times r$, $f\leftarrow Q.Pop()$
  \FOR{$o \in \Omega(f)$}
    \STATE $f'\leftarrow\mathcal{M}(f, \{o\})$
    \IF {$e^{\frac{c_{min} - \alpha(f')}{T}} > Rand()$}
      \STATE $Q$.Push($f'$)
    \ENDIF
    \IF {$ c_{min} >\alpha(f')$}
      \STATE $c_{min} \leftarrow \alpha(f')$, $f_{min} \leftarrow f'$
    \ENDIF
  \ENDFOR
\ENDWHILE

\STATE {\bfseries Return} The nearest ancestor of $f_{min}$ in $\mathcal{H}$, the operation sequence to reach $f_{min}$
\end{algorithmic}
\end{algorithm}

As shown in Algorithm~\ref{algo_max}, the algorithm takes minimum temperature $T_{low}$,
temperature decreasing rate $r$ for simulated annealing,
and search history $\mathcal{H}$ described in Section~\ref{sec_problem} as the input.
It outputs a neural architecture $f\in\mathcal{H}$ and a sequence of operations $O$ to morph $f$ into the new architecture.
From line 2 to 6, the searched architectures are pushed into the priority queue,
which sorts the elements according to the cost function value or the acquisition function value.
Since UCB is chosen as the acquisiton function,
$\alpha(f)$ is directly comparable with the history observation values $c^{(i)}$.
% $s[f]$ records which history architecture is $f$ morphed from.
% $O[f]$ records what operations are applied to morph $s[f]$ to $f$.
From line 7 to 18, it is the loop optimizing the acquisition function.
Following the setting in A* search,
in each iteration,
the architecture with the lowest acquisition function value is popped out to be expanded on line 8 to 10,
where $\Omega(f)$ is all the possible operations to morph the architecture $f$,
$\mathcal{M}(f, {o})$ is the function to morph the architecture $f$ with the operation sequence ${o}$.
However, not all the children are pushed into the priority queue for exploration purpose.
The decision of whether it is pushed into the queue is made by simulated annealing on line 11,
where $e^{\frac{c_{min} - \alpha(f')}{T}}$ is a typical acceptance function in simulated annealing.
$c_{min}$ and $f_{min}$ are updated from line 14 to 16, which record the minimum acquisition function value and the corresponding architecture.
\subsection{Graph-Level Network Morphism}
\label{sec_nm}

The third challenge is to maintain the intermediate output tensor shape consistency when morphing the architectures.
Previous work showed how to preserve the functionality of the layers the operators applied on,
namely layer-level morphism.
However, from a graph-level view, any change of a single layer could have a butterfly effect on the entire network.
Otherwise, it would break the input and output tensor shape consistency.
To tackle the challenge,
a graph-level morphism is proposed to find and morph the layers influenced by a layer-level operation in the entire network.

Follow the four network morphism operations on a neural network $f \in \mathcal{F}$ defined in~\cite{elsken2017simple},
which can all be reflected in the change of the computational graph $G$.
The first operation is inserting a layer to $f$ to make it deeper denoted as $deep(G, u)$,
where $u$ is the node marking the place to insert the layer.
The second one is widening a node in $f$ denoted as $wide(G, u)$,
where $u$ is the node representing the intermediate output tensor to be widened.
Widen here could be either making the output vector of the previous fully-connected layer of $u$ longer,
or adding more filters to the previous convolutional layer of $u$, depending on the type of the previous layer.
The third is adding an additive connection from node $u$ to node $v$ denoted as $add(G, u, v)$.
The fourth is adding an concatenative connection from node $u$ to node $v$ denoted as $concat(G, u, v)$.
For $deep(G, u)$, no other operation is needed except for initializing the weights of the newly added layer.
However, for all other three operations, more changes are required to $G$.

First, we define an effective area of $wide(G, u_0)$ as $\gamma$ to better describe where to change in the network.
The effective area is a set of nodes in the computational graph,
which can be recursively defined by the following rules:
1. $u_0 \in \gamma$.
2. $v\in \gamma$, if $\exists e_{u\rightarrow v} \not\in L_s$, $u\in\gamma$.
3. $v\in \gamma$, if $\exists e_{v\rightarrow u} \not\in L_s$, $u\in\gamma$.
$L_s$ is the set of fully-connected layers and convolutional layers.
Operation $wide(G, u_0)$ needs to change two set of layers, the previous layer set $L_p=\{e_{u\rightarrow v} \in L_s | v\in\gamma\}$,
which needs to output a wider tensor, and next layer set $L_n=\{e_{u\rightarrow v} \in L_s | u\in\gamma\}$,
which needs to input a wider tensor.
Second, for operator $add(G, u_0, v_0)$, additional pooling layers may be needed on the skip-connection.
$u_0$ and $v_0$ have the same number of channels, but their shape may differ because of the pooling layers between them.
So we need a set of pooling layers whose effect is the same as the combination of all the pooling layers between $u_0$ and $v_0$,
which is defined as $L_o=\{e\in L_{pool}|e\in p_{u_0\rightarrow v_0}\}$.
where $p_{u_0\rightarrow v_0}$ could be any path between $u_0$ and $v_0$,
$L_{pool}$ is the pooling layer set.
Another layer $L_c$ is used after to pooling layers to process $u_0$ to the same width as $v_0$.
Third, in $concat(G, u_0, v_0)$, the concatenated tensor is wider than the original tensor $v_0$.
The concatenated tensor is input to a new layer $L_c$ to reduce the width back to the same width as $v_0$.
Additional pooling layers are also needed for the concatenative connection.
% Third, the effect area of $concat(G, u_0, v_0)$ can be similarly defined by the following rules:
% 1. $u_0 \in \gamma$.
% 2. $v_0 \in \gamma$.
% 3. $v\in \gamma$, if $\exists e_{u\rightarrow v} \not\in L_s$, $u\in\gamma$.
% 4. $v\in \gamma$, if $\exists e_{v\rightarrow u} \not\in L_s$, $u\in\gamma \land u\neq u_0 \land u\neq v_0$.
% \noindent The $L_p$ and $L_n$ is the same as defined in the wide operation.

\subsection{Time Complexity Analysis}
As described at the start of Section 3,
Bayesian optimization can be roughly divided into three steps: update, generation, and observation.
The bottleneck of the algorithm efficiency is observation,
which involves the training of the generated neural architecture.
Let $n$ be the number of architectures in the search history.
The time complexity of the update is $O(n^2\log_2n)$.
In each generation, the kernel is computed between the new architectures during optimizing acquisition function and the ones in the search history,
the number of values in which is $O(nm)$,
where $m$ is the number of architectures computed during the optimization of the acquisition function.
The time complexity for computing $d(\cdot,\cdot)$ once is $O(l^2+s^3)$,
where $l$ and $s$ are the number of layers and skip-connections.
So the overall time complexity is $O(nm(l^2+s^3) + n^2\log_2n)$.
The magnitude of these factors is within the scope of tens.
So the time consumption of update and generation is trivial comparing to the observation.

\section{Auto-Keras}

Based on the proposed neural architecture search method,
we developed an open-source AutoML system, namely Auto-Keras.
% in which Keras \cite{chollet2015keras} is used for the construction and training of the neural networks.
It is named after Keras~\cite{chollet2015keras}, which is known for its simplicity in creating neural networks.
Similar to SMAC~\cite{hutter2011sequential}, TPOT~\cite{OlsonGECCO2016}, Auto-WEKA~\cite{thornton2013auto}, and Auto-Sklearn~\cite{feurer2015efficient},
the goal is to enable domain experts who are not familiar with machine learning technologies to use machine learning techniques easily.
However, Auto-Keras is focusing on the deep learning tasks, which is different from the systems focusing on the shallow models mentioned above.

Although, there are several AutoML services available on large cloud computing platforms,
three things are prohibiting the users from using them.
First, the cloud services are not free to use, which may not be affordable for everyone who wants to use AutoML techniques.
Second, the cloud-based AutoML usually requires complicated configurations of Docker containers and Kubernetes,
which is not easy for people without a rich computer science background.
Third, the AutoML service providers are honest-but-curious~\cite{chai2012verifiable}, which cannot guarantee the security and privacy of the data.
An open-source software, which is easily downloadable and runs locally, would solve these problems and make the AutoML accessible to everyone.
To bridge the gap, we developed Auto-Keras.

It is challenging, to design an easy-to-use and locally deployable system.
First, we need a concise and configurable application programming interface (API). For the users who don't have rich experience in programming, they could easily learn how to use the API.
For the advanced users, they can still configure the details of the system to meet their requirements.
Second, the local computation resources may be limited.
We need to make full use of the local computation resources to speed up the search.
Third, the available GPU memory may be of different sizes in different environments.
We need to adapt the neural architecture sizes to the GPU memory during the search.

\subsection{System Overview}

\begin{figure}
    \includegraphics[width=0.48\textwidth]{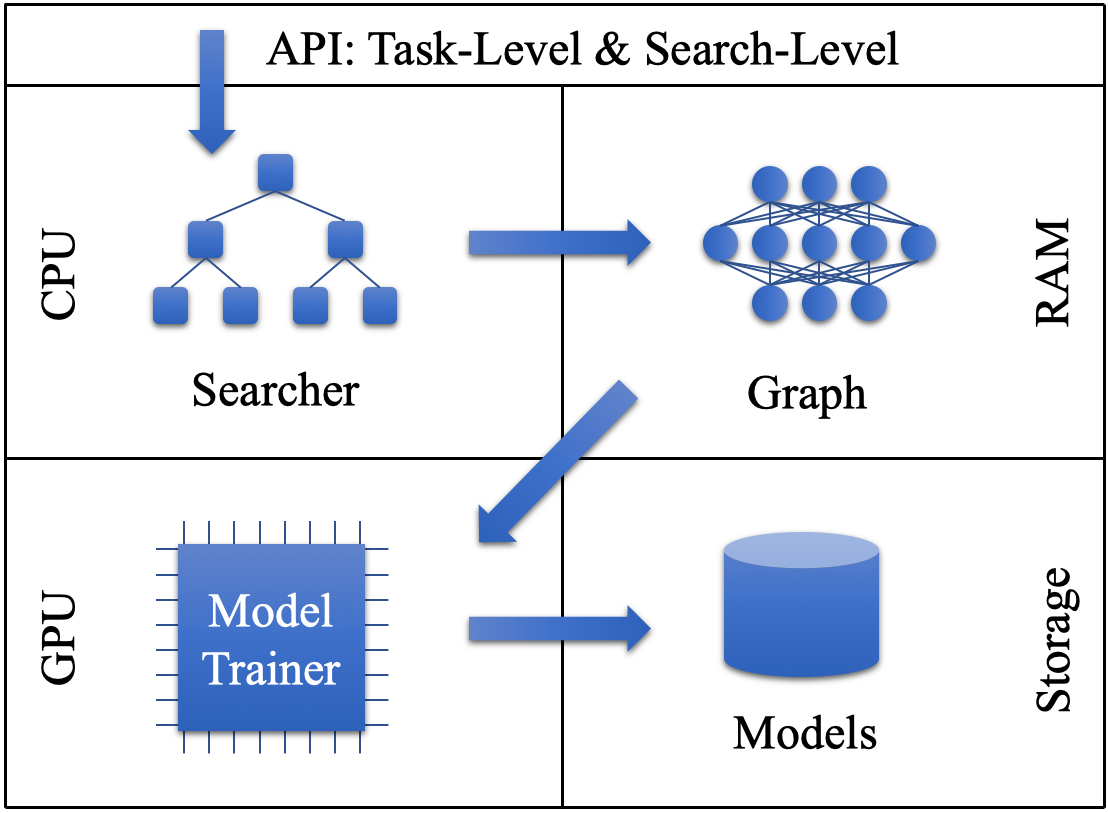}
    \caption{Auto-Keras System Overview. (1) User calls the API. (2) The Searcher generates neural architectures on CPU. (3) Graph builds real neural networks with parameters on RAM from the neural architectures. (4) The neural network is copied to GPU for training. (5) Trained neural networks are saved on storage devices.}
    \label{fig_system}
\end{figure}%

The system architecture of Auto-Keras is shown in Figure~\ref{fig_system}.
We design this architecture to fully make use of the computational resource of both CPU and GPU,
and utilize the memory efficiently by only placing the currently useful information on the RAM,
and save the rest on the storage devices, \textit{e.g.}, hard drives.
The top part is the API,
which is directly called by the users.
It is responsible for calling corresponding middle-level modules to complete certain functionalities.
% The middle part of the system consists of three modules.
The Searcher is the module of the neural architecture search algorithm containing Bayesian Optimizer and Gaussian Process.
These search algorithms run on CPU.
The Model Trainer is a module responsible for the computation on GPUs.
It trains given neural networks with the training data in a separate process for parallelism.
The Graph is the module processing the computational graphs of neural networks,
which is controlled by the Searcher for the network morphism operations.
The current neural architecture in the Graph is placed on RAM for faster access.
The Model Storage is a pool of trained models.
Since the size of the neural networks are large and cannot be stored all in memory,
the model storage saves all the trained models on the storage devices.
% It is capable of various training techniques to improve the final performance of the neural network including data augmentation and automated identification of convergence.

A typical workflow for the Auto-Keras system is as follows.
The user initiated a search for the best neural architecture for the dataset.
The API received the call, preprocess the dataset, and pass it to the Searcher to start the search.
The Bayesian Optimizer in the Searcher would generate a new architecture using CPU.
It calls the Graph module to build the generated neural architecture into a real neural network in the RAM.
The new neural architecture is copied the GPU for the Model Trainer to train with the dataset.
The trained model is saved in the Model Storage.
The performance of the model is feedback to the Searcher to update the Gaussian Process.

\subsection{Application Programming Interface}
The design of the API follows the classic design of the Scikit-Learn API~\cite{sklearn_api,scikit-learn},
which is concise and configurable.
The training of a neural network requires as few as three lines of code calling the constructor, the fit and predict function respectively.
% Most of the hyperparameters, e.g., number of epochs, data augmentation, are configurable through the API.
To accommodate the needs of different users, we designed two levels of APIs.
The first level is named as task-level.
The users only need to know their task, \textit{e.g.}, Image Classification, Text Regression, to use the API.
The second level is named search-level, which is for advanced users.
The user can search for a specific type of neural network architectures, \textit{e.g.}, multi-layer perceptron, convolutional neural network.
To use this API, they need to preprocess the dataset by themselves and know which type of neural network, \textit{e.g.}, CNN or MLP, is the best for their task.

Several accommodations have been implemented to enhance the user experience with the Auto-Keras package.
First, the user can restore and continue a previous search which might be accidentally killed.
From the users' perspective, the main difference of using Auto-Keras comparing with the AutoML systems aiming at shallow models is the much longer time consumption,
since a number of deep neural networks are trained during the neural architecture search.
It is possible for some accident to happen to kill the process before the search finishes.
Therefore, the search outputs all the searched neural network architectures with their trained parameters into a specific directory on the disk.
As long as the path to the directory is provided, the previous search can be restored.
% Second, all the searched architectures are visualized in the saved directory as PNG files.
Second, the user can export the search results, which are neural architectures, as saved Keras models for other usages.
Third, for advanced users,
they can specify all kinds of hyperparameters of the search process
and neural network optimization process by the default parameters in the interface.

\subsection{CPU and GPU Parallelism}
\begin{figure}
    \includegraphics[width=0.48\textwidth]{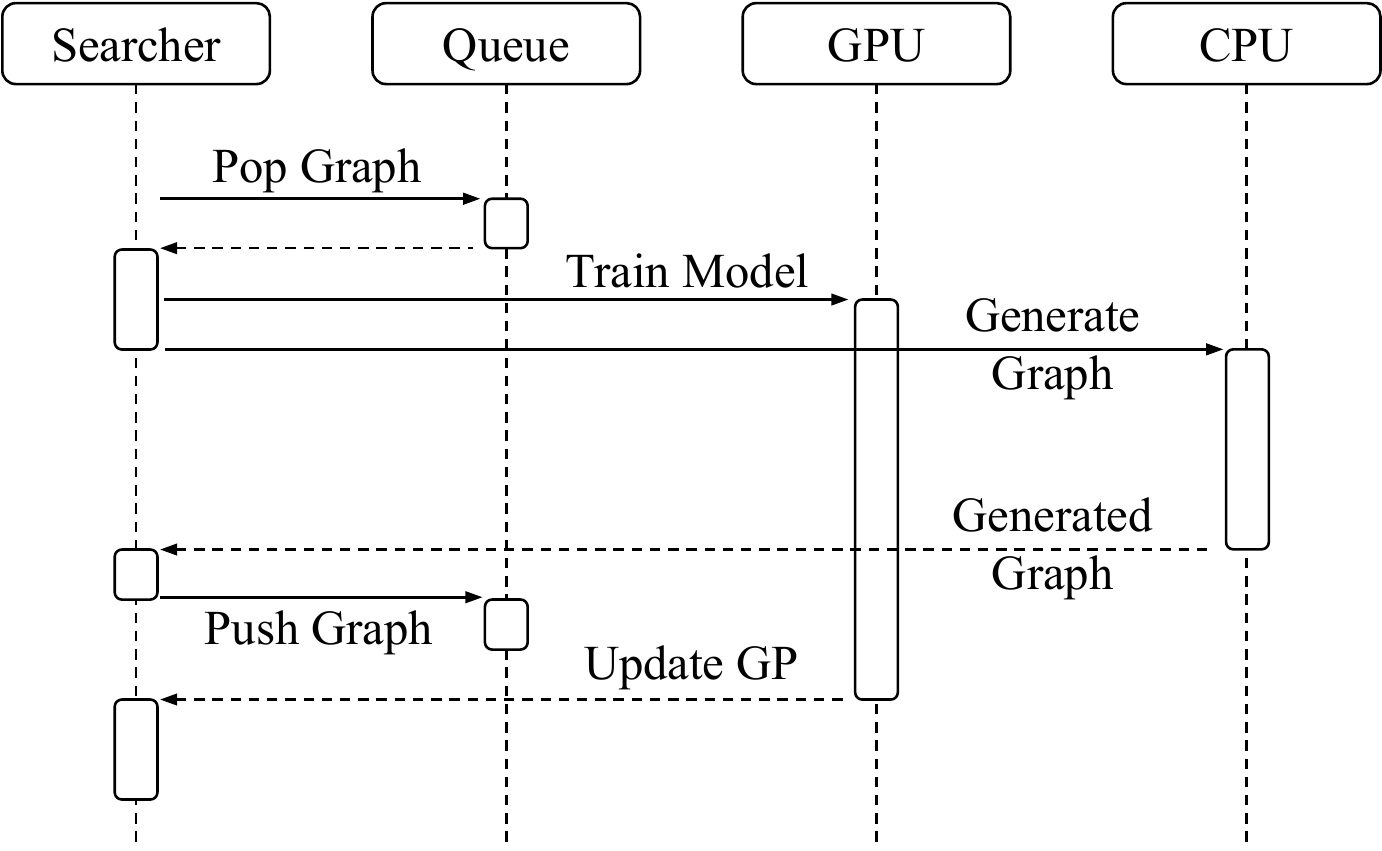}
    \caption{CPU and GPU Parallelism. During training the current neural architecture on GPU, CPU generates the next neural architecture (graph).}
    \label{fig_seq}
\end{figure}%

To make full use of the limited local computation resources, the program can run in parallel on the GPU and the CPU at the same time.
% It relies on the inner parallel mechanism of Keras to run across multiple GPUs during the training of the neural networks.
% The functional programming paradigm in python enables the rest of the computation to run in parallel across multiple CPUs.
If we do the observation (training of the current neural network), update, and generation of Bayesian optimization in sequential order.
The GPUs will be idle during the update and generation.
The CPUs will be idle during the observation.
To improve the efficiency, the observation is run in parallel with the generation in separated processes.
A training queue is maintained as a buffer for the Model Trainer.
Figure~\ref{fig_seq} shows the Sequence diagram of the parallelism between the CPU and the GPU.
First, the Searcher requests the queue to pop out a new graph and pass it to GPU to start training.
Second, while the GPU is busy, the searcher requests the CPU to generate a new graph.
At this time period, the GPU and the CPU work in parallel.
Third, the CPU returns the generated graph to the searcher, who pushes the graph into the queue.
Finally, the Model Trainer finished training the graph on the GPU and returns it to the Searcher to update the Gaussian process.
In this way, the idle time of GPU and CPU are dramatically reduced to improve the efficiency of the search process.

\subsection{GPU Memory Adaption}
Since different deployment environments have different limitations on the GPU memory usage,
the size of the neural networks needs to be limited according to the GPU memory.
Otherwise, the system would crash because of running out of GPU memory.
To tackle this challenge, we implement a memory estimation function on our own data structure for the neural architectures.
An integer value is used to mark the upper bound of the neural architecture size.
Any new computational graph whose estimated size exceeds the upper bound is discarded.
However, the system may still crash because the management of the GPU memory is very complicated, which cannot be precisely estimated.
So whenever it runs out of GPU memory, the upper bound is lowered down to further limit the size of the generated neural networks.

\section{Experiments}
In the experiments, we aim at answering the following questions.
1) How effective is the search algorithm with limited running time?
2) How much efficiency is gained from Bayesian optimization and network morphism?
3) What are the influences of the important hyperparameters of the search algorithm?
4) Does the proposed kernel function correctly measure the similarity among neural networks in terms of their actual performance?

\textbf{Datasets}
Three benchmark datasets, MNIST~\cite{lecun1998gradient}, CIFAR10~\cite{krizhevsky2009learning},
and FASHION~\cite{xiao2017/online} are used in the experiments to evaluate our method.
They prefer very different neural architectures to achieve good performance.
% MNIST and CIFAR10 are very famous image classification datasets.
% FASHION is the Fashion-MNIST, which is a dataset of outfits image classification.

\textbf{Baselines}
Four categories of baseline methods are used for comparison, which are elaborated as follows:
\begin{itemize}[leftmargin=*]
\item Straightforward Methods: random search (RAND) and grid search (GRID).
They search the number of convolutional layers and the width of those layers.
\item Conventional Methods: SPMT~\cite{snoek2012practical} and SMAC~\cite{hutter2011sequential}. Both SPMT and SMAC are designed for general hyperparameters tuning tasks of machine learning models instead of focusing on the deep neural networks.
They tune the 16 hyperparameters of a three-layer
convolutional neural network,
including the width, dropout rate, and regularization rate of each layer.
\item State-of-the-art Methods: SEAS~\cite{elsken2017simple}, NASBOT~\cite{kandasamy2018neural}. We carefully implemented the SEAS as described in their paper.
For NASBOT, since the experimental settings are very similar,
we directly trained their searched neural architecture in the paper. They did not search architectures for MNIST and FASHION dataset, so the results are omitted in our experiments.
\item Variants of the proposed method: BFS and BO. Our proposed method is denoted as AK. BFS replaces the Bayesian optimization in AK with the breadth-first search. BO is another variant, which does not employ network morphism to speed up the training. For AK, $\beta$ is set to 2.5, while $\lambda$ is set to 1 according to the parameter sensitivity analysis.
\end{itemize}
In addition, the performance of the deployed system of Auto-Keras (AK-DP) is also evaluated in the experiments. The difference from the AK above is that AK-DP uses various advanced techniques to improve the performance including learning rate scheduling, multiple manually defined initial architectures.

\textbf{Experimental Setting} The general experimental setting for evaluation is described as follows: First, the original training data of each dataset is further divided into training and validation sets by 80-20.
Second, the testing data of each dataset is used as the testing set.
Third, the initial architecture for SEAS, BO, BFS, and AK is a three-layer convolutional neural network with 64 filters in each layer.
Fourth, each method is run for 12 hours on a single GPU (NVIDIA GeForce GTX 1080 Ti) on the training and validation set with batch size of 64.
Fifth, the output architecture is trained with both the training and validation set.
Sixth, the testing set is used to evaluate the trained architecture.
Error rate is selected as the evaluation metric since all the datasets are for classification.
For a fair comparison, the same data processing and training procedures are used for all the methods.
The neural networks are trained for 200 epochs in all the experiments.
Notably, AK-DP uses a real deployed system setting, whose result is not directly comparable with the rest of the methods.
Except for AK-DP, all other methods are fairly compared using the same initial architecture to start the search.

\subsection{Evaluation of Effectiveness}
\label{sec_effective}

\begin{table}
  \caption{\label{tab_performance}Classification Error Rate}

\begin{tabular}{l c c c}
    \hline
    Methods & MNIST & CIFAR10 & FASHION \\
    \hline
    RANDOM & $1.79\%$ & $16.86\%$ & $11.36\%$ \\
    GRID & $1.68\%$  & $17.17\%$ & $10.28\%$\\
    \hline
    SPMT & $1.36\%$ & $14.68\%$ & $9.62\%$ \\
    SMAC & $1.43\%$ & $15.04\%$ & $10.87\%$ \\
    \hline
    SEAS & $1.07\%$ & $12.43\%$ & $8.05\%$ \\
    NASBOT & NA & $12.30\%$ & NA \\
    % ENAS & NA & $0.063$ & NA \\
    \hline
    BFS & $1.56\%$ & $13.84\%$ & $9.13\%$ \\
    BO & $1.83\%$ & $12.90\%$ & $7.99\%$ \\
    AK & $\boldsymbol{0.55\%}$ & $\boldsymbol{11.44\%}$ & $\boldsymbol{7.42\%}$ \\
    \hline
    AK-DP & $0.60\%$ & $3.60\%$ & $6.72\%$ \\
    \hline
   \end{tabular}
\end{table}

We first evaluate the effectiveness of the proposed method. The results are shown in Table~\ref{tab_performance}.
The following conclusions can be drawn based on the results.

(1)
AK-DP is evaluated to show the final performance of our system,
which shows deployed system (AK-DP) achieved state-of-the-art performance on all three datasets.

(2) The proposed method AK achieves the lowest error rate on all the three datasets,
which demonstrates that AK is able to find simple but effective architectures on small datasets (MNIST) and can explore more complicated structures on larger datasets (CIFAR10).

(3) The straightforward approaches and traditional approaches perform well on the MNIST dataset,
but poorly on the CIFAR10 dataset. This may come from the fact that: naive approaches like random search and grid search only try a limited number of architectures blindly while the two conventional approaches are unable to change the depth and skip-connections of the architectures.

%state of the art why
(4) Though the two state-of-the-art approaches achieve acceptable performance, SEAS could not beat our proposed model due to its subpar search strategy. The hill-climbing strategy it adopts only takes one step at each time in morphing the current best architecture, and the search tree structure is constrained to be unidirectionally extending. Comparatively speaking, NASBOT possesses stronger search expandability and also uses Bayesian optimization as our proposed method.
However, the low efficiency in training the neural architectures constrains its power in achieving comparable performance within a short time period. By contrast, the network morphism scheme along with the novel searching strategy ensures our model to achieve desirable performance with limited hardware resources and time budges.

(5) For the two variants of AK,
BFS preferentially considers searching a vast number of neighbors surrounding the initial architecture,
which constrains its power in reaching the better architectures away from the initialization. By comparison, BO can jump far from the initial architecture.
But without network morphism, it needs to train each neural architecture with much longer time, which limits the number of architectures it can search within a given time.

\subsection{Evaluation of Efficiency}
In this experiment, we try to evaluate the efficiency gain of the proposed method in two aspects.
First, we evaluate whether Bayesian optimization can really find better solutions with a limited number of observations.
Second, we evaluated whether network morphism can enhance the training efficiency.

\begin{figure}
    \includegraphics[width=0.48\textwidth]{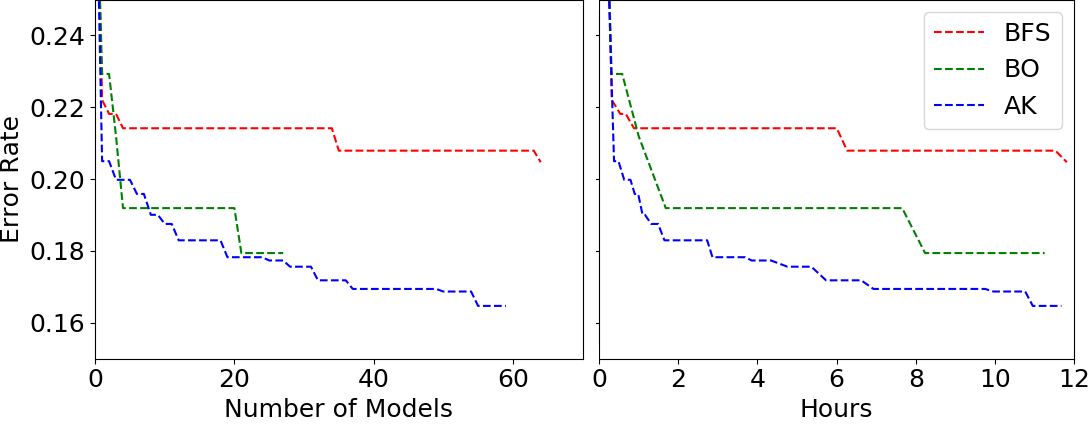}

    \caption{Evaluation of Efficiency. The two figures plot the same result with different X-axis.
    BFS uses network morphism.
    BO uses Bayesian optimization. AK uses both.}
    \label{fig_eff}
\end{figure}

We compare the proposed method AK with its two variants, BFS and BO, to show the efficiency gains from Bayesian optimization and network morphism, respectively.
BFS does not adopt Bayesian optimization but only network morphism, and use breadth-first search to select the network morphism operations. BO does not employ network morphism but only Bayesian optimization.
Each of the three methods is run on CIFAR10 for twelve hours.
The left part of Figure~\ref{fig_eff} shows the relation between the lowest error rate achieved and the number of neural networks searched.
The right part of Figure~\ref{fig_eff} shows the relation between the lowest error rate achieved and the searching time.

Two conclusions can be drawn by comparing BFS and AK. First, Bayesian optimization can efficiently find better architectures with a limited number of observations.
When searched the same number of neural architectures, AK could achieve a much lower error rate than BFS. It demonstrates that
Bayesian optimization could effectively guide the search in the right direction,
which is much more efficient in finding good architectures than the naive BFS approach. Second, the overhead created by Bayesian optimization during the search is low.
In the left part of Figure~\ref{fig_eff}, it shows BFS and AK searched similar numbers of neural networks within twelve hours.
BFS is a naive search strategy, which does not consume much time during the search besides training the neural networks.
AK searched slightly less neural architectures than BFS because of higher time complexity.

Two conclusions can be drawn by comparing BO and AK.
First, network morphism does not negatively impact the search performance.
In the left part of Figure~\ref{fig_eff}, when BO and AK search a similar number of neural architectures, they achieve similar lowest error rates.
Second, network morphism increases the training efficiency, thus improve the performance.
As shown in left part of Figure~\ref{fig_eff}, AK could search much more architectures than BO within the same amount of time due to the adoption of network morphism.
Since network morphism does not degrade the search performance, searching more architectures results in finding better architectures.
This could also be confirmed in the right part of Figure~\ref{fig_eff}. At the end of the searching time, AK achieves lower error rate than BO.

\subsection{Parameter Sensitivity Analysis}
We now analyze the impacts of the two most important hyperparameters in our proposed method, \textit{i.e.},
% $\lambda$ in Equation~\eqref{eq_distance} and $\beta$ in Equation~\eqref{eq_acq}. $\lambda$ balances the distance of layers and skip connections in the kernel function, and
% $\beta$ is the weight of the variance in the acquisition function,
% which balances the exploration and exploitation of the search strategy.
$\beta$ in Equation~\eqref{eq_acq} balancing the exploration and exploitation of the search strategy,
and $\lambda$ in Equation~\eqref{eq_distance} balancing the distance of layers and skip connections.
For other hyperparameters, since $r$ and $T_{low}$ in Algorithm~\ref{algo_max} are just normal hyperparameters of simulated annealing instead of important parameters directly related to neural architecture search, we do not delve into them here.
In this experiment, we use the CIFAR10 dataset as an example.
The rest of the experimental setting follows the setting of Section~\ref{sec_effective}.

\begin{figure}
        \includegraphics[width=0.47\textwidth]{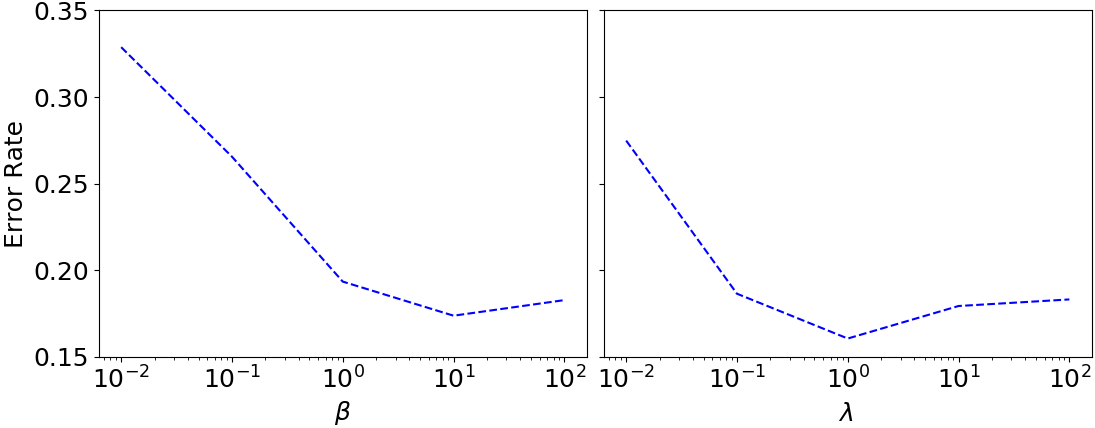}
    \caption{Parameter Sensitivity Analysis. $\beta$ balances the exploration and exploitation of the search strategy. $\lambda$ balances the distance of layers and skip connections.}
    \label{fig_param}
\end{figure}

From Figure~\ref{fig_param}, we can observe that the influences of $\beta$ and $\lambda$ to the performance of our method are similar.
As shown in the left part of Figure~\ref{fig_param}, with the increase of $\beta$ from $10^{-2}$ to $10^{2}$, the error rate decreases first and then increases.
If $\beta$ is too small, the search process is not explorative enough to search the architectures far from the initial architecture.
If it is too large, the search process would keep exploring the far points instead of trying the most promising architectures.
Similarly, as shown in the right part of Figure~\ref{fig_param}, the increase of $\lambda$ would downgrade the error rate at first and then upgrade it.
This is because if $\lambda$ is too small, the differences in the skip-connections of two neural architectures are ignored; conversely, if it is too large, the differences in the convolutional or fully-connected layers are ignored.
The differences in layers and skip-connections should be balanced in the kernel function for the entire framework to achieve a good performance.

\subsection{Evaluation of Kernel Quality}

\begin{figure}
    \begin{subfigure}{0.24\textwidth}
        \includegraphics[width=0.95\textwidth]{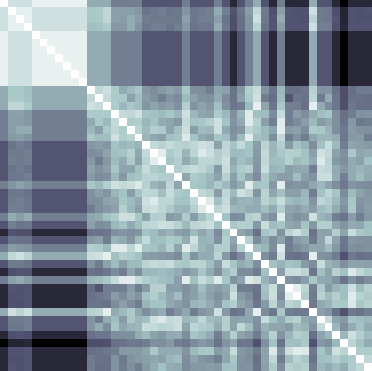}
        \caption{\label{fig_kernel1}Kernel Matrix}
    \end{subfigure}%
    \begin{subfigure}{0.24\textwidth}
        \includegraphics[width=0.95\textwidth]{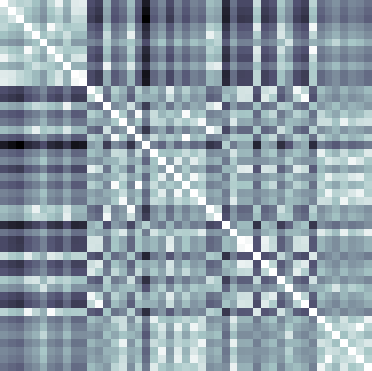}
        \caption{\label{fig_kernel2}Performance Similarity}
    \end{subfigure}
    \caption{\label{fig_kernel}Kernel and Performance Matrix Visualization. (a) shows the proposed kernel matrix. (b) is a matrix of similarity in the performance of the neural architectures.}
\end{figure}
To show the quality of the edit-distance neural network kernel,
we investigate the difference between the two matrices $\boldsymbol{K}$ and $\boldsymbol{P}$.
$\boldsymbol{K}_{n\times n}$ is the kernel matrix,
where $\boldsymbol{K}_{i,j}=\kappa(f^{(i)}, f^{(j)})$.
$\boldsymbol{P}_{n\times n}$ describes the similarity of the actual performance between neural networks,
where $\boldsymbol{P}_{i,j}=-|c^{(i)}-c^{(j)}|$,
$c^{(i)}$ is the cost function value in the search history $\mathcal{H}$ described in Section~\ref{sec_method}.
We use CIFAR10 as an example here, and adopt error rate as the cost metric.
Since the values in $\boldsymbol{K}$ and $\boldsymbol{P}$ are in different scales, both matrices are normalized to the range $[-1, 1]$.
We quantitatively measure the difference between $\boldsymbol{K}$ and $\boldsymbol{P}$ with mean square error,
which is $1.12\times10^{-1}$.

$\boldsymbol{K}$ and $\boldsymbol{P}$ are visualized in Figure~\ref{fig_kernel1} and~\ref{fig_kernel2}.
Lighter color means larger values.
There are two patterns can be observed in the figures.

First, the white diagonal of Figure~\ref{fig_kernel1} and~\ref{fig_kernel2}.
According to the definiteness property of the kernel, $\kappa(f_x, f_x)=1, \forall f_x \in \mathcal{F}$, thus
the diagonal of $\boldsymbol{K}$ is always 1.
It is the same for $\boldsymbol{P}$ since no difference exists in the performance of the same neural network.

Second, there is a small light square area on the upper left of Figure~\ref{fig_kernel1}.
These are the initial neural architectures to train the Bayesian optimizer, which are neighbors to each other in terms of network morphism operations.
A similar pattern is reflected in Figure~\ref{fig_kernel2}, which indicates that
when the kernel measures two architectures as similar,
they tend to have similar performance.

\section{Conclusion and Future Work}
In this paper, a novel method for efficient neural architecture search with network morphism is proposed.
It enables Bayesian optimization to guide the search by designing a neural network kernel,
and an algorithm for optimizing acquisition function in tree-structured space.
The proposed method is wrapped into an open-source AutoML system, namely Auto-Keras, which can be easily downloaded and used with an extremely simple interface.
The method has shown good performance in the experiments and outperformed several traditional hyperparameter-tuning methods and state-of-the-art neural architecture search methods.
We plan to study the following open questions in future work.
(1) The search space may be expanded to the recurrent neural networks.
(2) Tune the neural architecture and the hyperparameters of the training process jointly.
(3) Design task-oriented NAS to solve specific machine learning problems, \textit{e.g.}, image segmentation~\cite{liu2019auto} and object detection~\cite{liu2016ssd}.
\bibliographystyle{ACM-Reference-Format}
\bibliography{ref}

\appendix
\section*{Appendix: Reproducibility}
In this section, we provide the details of our implementation and proofs for reproducibility.
\begin{itemize}
  \item The default architectures used to initialized are introduced.
\item  The details of the implementation of the four network morphism operations are provided.
\item The details of preprocessing the datasets are shown.
\item  The details of the training process are described.
\item The proof of the validity of the kernel function is provided.
\item The process of using $\rho(\cdot)$ to distort the approximated edit-distance of the neural architectures $d(\cdot,\cdot)$ is introduced.
\end{itemize}
\textbf{Notably, the code and detailed documentation are available at Auto-Keras official website} (\textcolor{blue}{\url{https://autokeras.com}}).

\section{Default Architectures}
As we introduced in the experiment section, for all other methods except AK-DP,
are using the same three-layer convolutional neural network as the default architecture.
The AK-DP is initialized with ResNet, DenseNet and the three-layer CNN.
In the current implementation, ResNet18 and DenseNet121 specifically are chosen as the among all the ResNet and DenseNet architectures.

The three-layer CNN is constructed as follows.
Each convolutional layer is actually a convolutional block of a ReLU layer,
a batch-normalization layer, the convolutional layer, and a pooling layer.
All the convolutional layers are with kernel size equal to three, stride equal to one, and number of filters equal to 64.

All the default architectures share the same fully-connected layers design.
After all the convolutional layers, the output tensor passes through a global average pooling layer followed by a dropout layer, a fully-connected layer of 64 neurons, a ReLU layer, another fully-connected layer, and a softmax layer.

\section{Network Morphism Implementation}
The implementation of the network morphism is introduced from two aspects.
First, we describe how the new weights are initialized.
Second, we introduce a pool of possible operations which the Bayesian optimizer can select from, \textit{e.g.} the possible start and end points of a skip connection.

The four network morphism operations all involve adding new weights during inserting new layers and expanding existing layers.
We initialize the newly added weights with zeros.
However, it would create a symmetry prohibiting the newly added weights to learn different values during backpropagation.
We follow the Net2Net~\cite{chen2015net2net} to add noise to break the symmetry.
The amount of noise added is the largest noise possible not changing the output.

There are a large amount of possible network morphism operations we can choose.
Although there are only four types of operations we can choose,
a parameter of the operation can be set to a large number of different values.
For example, when we use the $deep(G, u)$ operation, we need to choose the location $u$ to insert the layer.
In the tree-structured search, we actually cannot exhaust all the operations to get all the children.
We will keep sampling from the possible operations until we reach eight children for a node.
For the sampling, we randomly sample an operation from $deep$, $wide$ and skip ($add$ and $concat$), with equally likely probability.
The parameters of the corresponding operation are sampled accordingly.
If it is the $deep$ operation, we need to decide the location to insert the layer.
In our implementation, any location except right after a skip-connection.
Moreover, we support inserting not only convolutional layers,
but activation layers, batch-normalization layers, dropout layer, and fully-connected layers as well.
They are randomly sampled with equally likely probability.
If it is the $wide$ operation, we need to choose the layer to be widened.
It can be any convolutional layer or fully-connected layer,
which are randomly sampled with equally likely probability.
If it is the skip operations, we need to decide if it is $add$ or $concat$.
The start point and end point of a skip-connection can be the output of any layer except the already-exist skip-connection layers.
So all the possible skip-connections are generated in the form of tuples of the start point, end point and type ($add$ or $concat$), among which we randomly sample a skip-connection with equally likely probability.

\section{Preprocessing the Datasets}
The benchmark datasets, \textit{e.g.}, MNIST, CIFAR10, FASHION, are preprocessed before the neural architecture search.
It involves normalization and data augmentation.
We normalize the data to the standard normal distribution.
For each channel, a mean and a standard deviation are calculated
since the values in different channels may have different distributions.
The mean and standard deviation are calculated using the training and validation set together.
The testing set is normalized using the same values.
The data augmentation includes random crop, random horizontal flip, and cutout,
which can improve the robustness of the trained model.

\section{Performance Estimation}
During the observation phase, we need to estimate the performance of a neural architecture to update the Gaussian process model in Bayesian optimization.
Since the quality of the observed performances of the neural architectures
is essential to the neural architecture search algorithm,
we propose to train the neural architectures
instead of using the performance estimation strategies used in literatures~\cite{brock2017smash,pham2018efficient,elsken2018neural}.
The quality of the observations is essential to the neural architecture search algorithm.
So the neural architectures are trained during the search in our proposed method.

There two important requirements for the training process.
First, it needs to be adaptive to different architectures.
Different neural networks require different numbers of epochs in training to converge.
Second, it should not be affected by the noise in the performance curve.
The final metric value, \textit{e.g.}, mean squared error or accuracy, on the validation set is not the best performance estimation
since there is random noise in it.

To be adaptive to architectures of different sizes,
we use the same strategy as the early stop criterion in the multi-layer perceptron algorithm in Scikit-Learn~\cite{scikit-learn}.
It sets a maximum threshold $\tau$.
If the loss of the validation set does not decrease in $\tau$ epochs, the training stops.
Comparing with the methods using a fixed number of training epochs,
it is more adaptive to different neural architectures.

To avoid being affected by the noise in the performance,
the mean of metric values of the last $\tau$ epochs on the validation set is used as the estimated performance for the given neural architecture.
It is more accurate than the final metric value on the validation set.

\section{Validity of the Kernel}
\label{sec_proof}
\textbf{Theorem 1.} $d(f_a, f_b)$ is a metric space distance.

\textbf{\textit{Proof of Theorem 1:}}

Theorem 1 is proved by proving the non-negativity, definiteness, symmetry,
and triangle inequality of $d$.

\textbf{Non-negativity}:

$\forall f_x\;f_y\in\mathcal{F}$, $d(f_x, f_y) \geq 0$.

From the definition of $w(l)$ in Equation~\eqref{eq_small_dl}, $\forall l$, $w(l) > 0$.
$\therefore \forall l_x \;l_y$, $d_l(l_x, l_y) \geq 0$.
$\therefore \forall L_x \;L_y$, $D_l(L_x, L_y) \geq 0$.
Similarly,
$\forall s_x \;s_y$, $d_s(s_x, s_y) \geq 0$,
and
$\forall S_x \;S_y$, $D_s(S_x, S_y) \geq 0$.
In conclusion, $\forall f_x\;f_y\in\mathcal{F}$, $d(f_x, f_y) \geq 0$.

\textbf{Definiteness}:

$f_a=f_b \iff d(f_a, f_b)=0$
.

$f_a=f_b \implies d(f_a, f_b)=0$ is trivial.
To prove $d(f_a, f_b)=0 \implies f_a=f_b$, let $d(f_a, f_b)=0$.
$\because \forall L_x \;L_y$, $D_l(L_x, L_y) \geq 0$
and $\forall S_x \;S_y$, $D_s(S_x, S_y) \geq 0$.
Let $L_a$ and $L_b$ be the layer sets of $f_a$ and $f_b$.
Let $S_a$ and $S_b$ be the skip-connection sets of $f_a$ and $f_b$.

$\therefore D_l(L_a, L_b)=0$ and $D_s(S_a, S_b)= 0$.
$\because \forall l_x \;l_y$, $d_l(l_x, l_y) \geq 0$
and $\forall s_x \;s_y$, $d_s(s_x, s_y) \geq 0$.
$\therefore |L_a|=|L_b|$,
$|S_a| =|S_b|$,
$\forall l_a\in L_a$, $l_b=\varphi_l(l_a)\in L_b$, $d_l(l_a, l_b)=0$,
$\forall s_a\in S_a$, $s_b=\varphi_s(s_a)\in S_b$, $d_s(s_a, s_b)=0$.
According to Equation~\eqref{eq_small_dl},
each of the layers in $f_a$ has the same width as the matched layer in $f_b$,
According to the restrictions of $\varphi_l(\cdot)$,
the matched layers are in the same order,
and all the layers are matched,
\textit{i.e.} the layers of the two networks are exactly the same.
Similarly,
the skip-connections in the two neural networks are exactly the same.
$\therefore f_a = f_b$.
So $d(f_a, f_b)=0 \implies f_a=f_b$, let $d(f_a, f_b)=0$.
Finally,
$f_a=f_b \iff d(f_a, f_b)$
.

\textbf{Symmetry}:

$\forall f_x\; f_y\in\mathcal{F}$, $d(f_x, f_y) = d(f_y, f_x)$.

Let $f_a$ and $f_b$ be two neural networks in $\mathcal{F}$,
Let $L_a$ and $L_b$ be the layer sets of $f_a$ and $f_b$.
If $|L_a|\neq|L_b|$, $D_l(L_a, L_b)=D_l(L_b, L_a)$
since it will always swap $L_a$ and $L_b$ if $L_a$ has more layers.
If $|L_a|=|L_b|$, $D_l(L_a, L_b)=D_l(L_b, L_a)$
since $\varphi_l(\cdot)$ is undirected, and $d_l(\cdot, \cdot)$ is symmetric.
Similarly, $D_s(\cdot, \cdot)$ is symmetric.
In conclusion, $\forall f_x\; f_y\in\mathcal{F}$, $d(f_x, f_y) = d(f_y, f_x)$.

\textbf{Triangle Inequality}:

$\forall f_x\;f_y\;f_z\in\mathcal{F}$, $d(f_x, f_y) \leq d(f_x, f_z) + d(f_z, f_y)$.

Let $l_x$, $l_y$, $l_z$ be neural network layers of any width.
If $w(l_x)<w(l_y)<w(l_z)$,
$d_l(l_x, l_y)=\frac{w(l_y) - w(l_x)}{w(l_y)}=2-\frac{w(l_x) + w(l_y)}{w(l_y)} \leq 2-\frac{w(l_x) + w(l_y)}{w(l_z)} = d_l(l_x, l_z) + d_l(l_z, l_y)$.
If $w(l_x)\leq w(l_z)\leq w(l_y)$,
$d_l(l_x, l_y)=\frac{w(l_y) - w(l_x)}{w(l_y)}=\frac{w(l_y) - w(l_z)}{w(l_y)}+\frac{w(l_z) - w(l_x)}{w(l_y)} \leq  \frac{w(l_y) - w(l_z)}{w(l_y)}+\frac{w(l_z) - w(l_x)}{w(l_z)} = d_l(l_x, l_z) + d_l(l_z, l_y)$.
If $w(l_z)\leq w(l_x)\leq w(l_y)$,
$d_l(l_x, l_y)=\frac{w(l_y) - w(l_x)}{w(l_y)}=2-\frac{w(l_y)}{w(l_y)}-\frac{w(l_x)}{w(l_y)} \leq 2-\frac{w(l_z)}{w(l_x)}-\frac{w(l_x)}{w(l_y)} \leq 2-\frac{w(l_z)}{w(l_x)} -\frac{w(l_z)}{w(l_y)} = d_l(l_x, l_z) + d_l(l_z, l_y)$.
By the symmetry property of $d_l(\cdot,\cdot)$, the rest of the orders of $w(l_x)$, $w(l_y)$ and $w(l_z)$ also satisfy the triangle inequality.
$\therefore \forall l_x\;l_y\;l_z$, $d_l(l_x, l_y) \leq d_l(l_x, l_z) + d_l(l_z, l_y)$.

$\forall L_a\;L_b\;L_c$, given $\varphi_{l:a\rightarrow c}$ and $\varphi_{l:c\rightarrow b}$ used to compute $D_l(L_a, $ $L_c)$ and $D_l(L_c, L_b)$,
we are able to construct $\varphi_{l:a\rightarrow b}$ to compute $D_l(L_a, L_b)$ satisfies $D_l(L_a, L_b)\leq D_l(L_a, L_c) +D_l(L_c, L_b)$.

Let $L_{a1} = \{\;l\;|\;\varphi_{l:a\rightarrow c}(l)\neq\varnothing\;\land\;\varphi_{l:c\rightarrow b}(\varphi_{l:c\rightarrow a}(l))\neq\varnothing\}$.
$L_{b1} = \{\;l\;|\;l=\varphi_{l:c\rightarrow b}(\varphi_{l:a\rightarrow c}(l')),\;l'\in L_{a1}\}$,
$L_{c1} = \{\;l\;|\;l=\varphi_{l:a\rightarrow c}(l')\neq\varnothing,\;l'\in L_{a1}\}$,
$L_{a2} = L_a - L_{a1}$,
$L_{b2} = L_b - L_{b1}$,
$L_{c2} = L_c - L_{c1}$.

From the definition of $D_l(\cdot, \cdot)$, with the current matching functions $\varphi_{l:a\rightarrow c}$ and $\varphi_{l:c\rightarrow b}$,
$D_l(L_a, L_c)=$ $D_l(L_{a1},$ $L_{c1})+$ $D_l$ $(L_{a2}$ $,L_{c2})$ and
$D_l(L_c, L_b)=$ $D_l(L_{c1},$ $L_{b1})+$ $D_l(L_{c2}$ $,L_{b2})$.
First, $\forall l_a \in L_{a1}$ is matched to $l_b=\varphi_{l:c\rightarrow b}(\varphi_{l:a\rightarrow c}(l_a))\in L_b$.
Since the triangle inequality property of $d_l(\cdot,\cdot)$,
$D_l(L_{a1}, L_{b1})\leq$ $D_l(L_{a1},$ $L_{c1})+$ $D_l(L_{c1},$ $L_{b1})$.
Second, the rest of the $l_a\in L_a$ and $l_b\in L_b$ are free to match with each other.

Let $L_{a21} = \{\;l\;|\;\varphi_{l:a\rightarrow c}(l)\neq\varnothing\;\land\;\varphi_{l:c\rightarrow b}(\varphi_{l:c\rightarrow a}(l))=\varnothing\}$,
$L_{b21} = \{\;l\;|\;l=\varphi_{l:c\rightarrow b}(l')\neq\varnothing,\;l'\in L_{c2}\}$,
$L_{c21} = \{\;l\;|\;l=\varphi_{l:a\rightarrow c}(l')\neq\varnothing,\;l'\in L_{a2}\}$,
$L_{a22} = L_{a2} - L_{a21}$,
$L_{b22} = L_{b2} - L_{b21}$,
$L_{c22} = L_{c2} - L_{c21}$.

From the definition of $D_l(\cdot, \cdot)$, with the current matching functions $\varphi_{l:a\rightarrow c}$ and $\varphi_{l:c\rightarrow b}$,
$D_l(L_{a2}, L_{c2})=D_l(L_{a21}, L_{c21})$ $+D_l$ $(L_{a22},$ $L_{c22})$
and $D_l(L_{c2}, L_{b2})=D_l(L_{c22}, L_{b21}) +D_l(L_{c21},$ $L_{b22})$.

$\because D_l(L_{a22}, L_{c22}) + D_l(L_{c21}, L_{b22}) \geq |L_{a2}|$

and
$D_l(L_{a21},$ $L_{c21})$ $+ D_l(L_{c22}, L_{b21}) \geq |L_{b2}|$

$\therefore D_l(L_{a2}, L_{b2})\leq |L_{a2}| + |L_{b2}|$ $\leq D_l(L_{a2}, L_{c2}) +D_l(L_{c2}, L_{b2})$.

So $D_l(L_a, L_b)\leq D_l(L_a, L_c) +D_l(L_c, L_b)$.

Similarly, $D_s(S_a, S_b)\leq D_s(S_a, S_c) +D_s(S_c, S_b)$.

Finally, $\forall f_x\;f_y\;f_z\in\mathcal{F}$, $d(f_x, f_y) \leq d(f_x, f_z) + d(f_z, f_y)$.

In conclusion, $d(f_a, f_b)$ is a metric space distance.
\hfill$\square$

\textbf{Theorem 2.} $\kappa(f_a, f_b)$ is a valid kernel.

\textbf{\textit{Proof of Theorem 2:}}
The kernel matrix of generalized RBF kernel in the form of $e^{-\gamma D^2(x, y)}$ is positive definite
if and only if there is an isometric embedding in Euclidean space for the metric space with metric $D$~\cite{haasdonk2004learning}.
Any finite metric space distance can be isometrically embedded into Euclidean space by changing the scale of the distance measurement~\cite{maehara2013euclidean}.
By using Bourgain theorem~\cite{bourgain1985lipschitz},
metric space $d$ is embedded to Euclidean space with distortion.
$\rho(d(f_a, f_b))$ is the embedded distance for $d(f_a, f_b)$.
Therefore, $e^{-\rho^2(d(f_a, f_b))}$ is always positive definite.
So $\kappa(f_a, f_b)$ is a valid kernel.\hfill$\square$

\section{Distance Distortion}
In this section, we introduce how Bourgain theorem is used
to distort the learned calculated edit-distance into an isometrically embeddable distance for Euclidean space in the Bayesian optimization process.

From Bourgain theorem, a Bourgain embedding algorithm is designed.
The input for the algorithm is a metric distance matrix.
Here we use the edit-distance matrix of neural architectures.
The outputs of the algorithm are some vectors in Euclidean space corresponding to the instances.
In our case, the instances are neural architectures.
From these vectors, we can calculate a new distance matrix using Euclidean distance.
The objective of calculating these vectors is to minimize the difference between the new distance matrix and the input distance matrix, \textit{i.e.}, minimize the distortions on the distances.

We apply this Bourgain algorithm during the update process of the Bayesian optimization.
The edit-distance matrix of previous training examples, \textit{i.e.}, the neural architectures, is stored in memory.
Whenever new examples are used to train the Bayesian optimization,
the edit-distance is expanded to include the new distances.
The distorted distance matrix is computed using Bourgain algorithm from the expanded edit-distance matrix.
It is isometrically embeddable to the Euclidean space.
The kernel matrix computed using the distorted distance matrix is a valid kernel.

\end{document}